\documentclass[letterpaper, 10 pt, conference]{ieeeconf}  

\IEEEoverridecommandlockouts                              

\usepackage{times}

\usepackage{tabularx}
\usepackage{booktabs}
\usepackage{graphicx}
\usepackage{multirow}
\usepackage{censor}
\usepackage{amssymb}
\usepackage{array}
\usepackage{amsmath}
\usepackage{cleveref}
\usepackage{cite}
\crefname{figure}{Fig.}{Figs.}
\Crefname{figure}{Fig.}{Figs.}

\crefname{table}{Table}{Tables}
\Crefname{table}{Table}{Tables}

\crefname{section}{Sec.}{Secs.}
\Crefname{section}{Sec.}{Secs.}

\newcolumntype{Y}{>{\centering\arraybackslash}X}

\title{\LARGE \bf
H-WM: Robotic Task and Motion Planning Guided by Hierarchical World Model
}

\author{
Jinbang Huang$^{1*}$,
Wenyuan Chen$^{1,2*}$,
Zhiyuan Li$^{1,2}$,
Oscar Pang$^{1,2}$,
Xiao Hu$^{1}$,
Lingfeng Zhang$^{1}$,
Yuanzhao Hu $^{1,3}$,\\
{Mark Coates}$^{4}$,
{Tongtong Cao}$^{1}$,
{Xingyue Quan}$^{1}$,
{Zhanguang Zhang}$^{1}$,
{Yingxue Zhang}$^{1}$%
\thanks{\scriptsize $^{1}$Huawei Noah’s Ark Lab, $^{2}$University of Toronto, $^{3}$ University of British Columbia, $^{4}$ McGill University, $^{*}$Equal contribution, Correspond to: jinbang.huang.work@gmail.com, \{zhanguang.zhang, yingxue.zhang\}@huawei.com}
}

\begin{document}

\maketitle
\thispagestyle{empty}
\pagestyle{empty}

\begin{abstract}

World models are becoming central to robotic planning and control by predicting future state transitions. Existing approaches mainly rely on visual, latent, or language prediction, which can be difficult to ground in executable robot actions and prone to compounding errors over long horizons. In contrast, traditional robotic task and motion planning enables structured long-horizon reasoning through compact symbolic representations of world transitions, but typically lacks synchronized visual prediction.
We propose Hierarchical World Model (H-WM), which jointly predicts logical and visual state transitions by combining a high-level logical world model with a low-level visual world model. The predicted logical actions and latent visual state transitions are jointly incorporated into Vision–Language–Action (VLA) models as intermediate state guidance for long-horizon task execution. Experiments on three long-horizon benchmarks and real robots show that H-WM consistently improves VLA's performance by stabilizing long-horizon execution and mitigating error accumulation. We also construct LIBERO-Logic, a frame-level aligned dataset that pairs visual observations and continuous robot states with logical actions and predicate-based logical states.

\end{abstract}


\section{Introduction}

Recent advances in VLA models have enabled robotic systems that tightly couple multi-modal perception and control via large pre-trained foundation models, achieving stronger generalization than traditional modular pipelines. However, most existing VLA methods adopt an end-to-end paradigm that maps visual observations and language instructions directly to low-level actions, resulting in degraded performance on long-horizon tasks~\cite{yang2025lohovla}. This failure is driven by compounding execution errors, ambiguous goal specifications, limited intermediate supervision, and overfitting to task-specific representations.

A natural response to these limitations is to introduce richer intermediate guidance; however, existing approaches fall into three dominant paradigms, each with unsolved shortcomings. First, LLM-based hierarchical planners decompose tasks into subgoals or action sequences~\cite{shi2025}, but are fundamentally constrained by language as the intermediate interface: LLMs struggle to reason about physical constraints, and their vague, unstructured representations lead to semantic–execution misalignment~\cite{shi2025,chengoal}. Second, world-model-based approaches aim to provide predictive visual guidance~\cite{shao2025,xiang2025pan}, yet existing formulations suffer from long-horizon settings where compounding prediction errors degrade planning reliability. Third, classical Task and Motion Planning (TAMP) achieves long-horizon consistency through explicit logical world models for symbolic reasoning~\cite{Kaelbling2011-mz}, but relies on manually designed abstractions and engineered symbolic representations that are weakly aligned with raw visual observations, resulting in brittleness to perception noise and difficulty in scaling to unstructured environments~\cite{Silver2021-mv}. Consequently, none of these paradigms delivers the informative, grounded, and long-horizon–robust guidance required for reliable VLA execution.

In this paper, we propose a novel hierarchical world model (H-WM) that couples long-horizon symbolic reasoning with perceptually grounded visual prediction to provide intermediate guidance for VLA models. First, a logical world model predicts structured action--state transitions and searches over candidate symbolic trajectories, promoting logically consistent task progression through trajectory evaluation and explicit goal-progress scoring. Second, a latent visual world model conditions on each selected logical transition and predicts the terminal visual representation of the corresponding subtask, avoiding autoregressive visual rollouts. The predicted logical action and visual subgoal are then jointly injected into a guided VLA for continuous low-level execution. Together, H-WM bridges high-level task reasoning and low-level perceptual control without requiring dense pixel-level prediction over the full task horizon. To summarize, our key contributions are: \textbf{(1)} A hierarchical world-modeling framework that combines symbolic transitions with visual subgoal prediction to ground robot execution.
\textbf{(2)} A logic-conditioned latent visual world model that directly predicts the visual representation of a subtask's terminal state from the predicted logical transition.
\textbf{(3)} A guided VLA architecture that jointly incorporates logical actions and predicted visual subgoals for continuous low-level execution.
\textbf{(4)} LIBERO-Logic, a robot manipulation dataset providing frame-level alignment among RGB observations, continuous robot states, logical actions, and logical states.

\begin{figure*}[t]
    \centering
    \includegraphics[width=\textwidth]{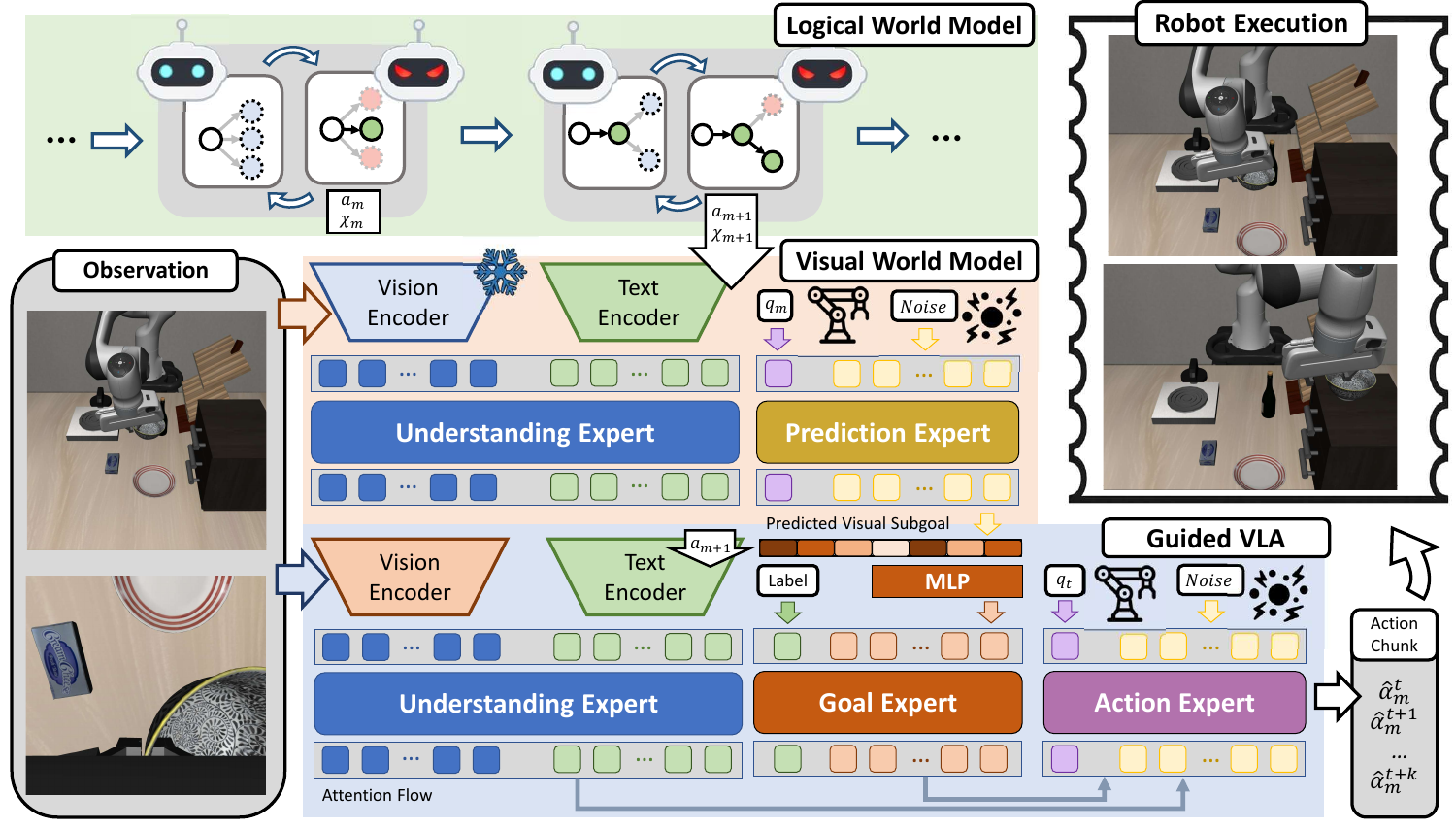}
\caption{\textbf{Overview of H-WM and guided VLA execution.} H-WM jointly models logical and visual transitions at the subtask level \(m\), while the VLA performs low-level control at the higher-frequency time step \(t\). The logical world model searches and evaluates candidate transitions to predict the next logical action \(a_{m+1}\) and logical state \(\mathcal{X}_{m+1}\). Conditioned on the current observation \(obs_m\), robot configuration \(q_m\), \(a_{m+1}\), and \(\mathcal{X}_{m+1}\), the visual world model uses an understanding expert and a prediction expert to generate the latent visual subgoal \(f_{\mathrm{pred}}^{m+1}\). During execution, the understanding expert encodes \(obs_t\) and \(a_{m+1}\), while the goal expert processes \(f_{\mathrm{pred}}^{m+1}\). The action expert takes \(q_t\) and noisy action tokens, attends to both representations, and generates the low-level action chunk \(\hat{\alpha}_{m}^{t:t+k}\), grounding long-horizon logical guidance into visually conditioned robot execution.}
    \label{fig:Framework}
    \vspace{-4mm}
\end{figure*}

\section{Related Work}

\subsection{World Model}

World models are powerful paradigms for robotics. Recent work has shown their effectiveness across a range of robotic settings, including policy learning~\cite{hafner2025training,li2025unified}, policy evaluation~\cite{quevedo2025evaluating,li2025worldeval}, test-time rollout simulation~\cite{bar2025navigation,yang2025mindjourney}, and unified model–policy training~\cite{zhao2025cot,bi2025motus,zhang2025dreamvla}. World models enable reliable future reasoning and can generate large-scale multi-modal data~\cite{zhao2025cot,bi2025motus,zhang2025dreamvla}. Recent systems further demonstrate that world models can be directly adapted for action generation~\cite{kim2026cosmos}.
Most existing approaches focus on pixel-level world modeling, yet real-world dynamics can be captured at multiple abstraction levels. Lower-level representations offer high expressivity but suffer from poor sample efficiency and generalization~\cite{xie2019improvisation,hoque2021visuo}, whereas higher-level abstractions exhibit the opposite trade-off~\cite{manuelli2020keypoints,driess2023}. Achieving a balance across levels remains challenging~\cite{xing2025critiques}.

\subsection{Vision–Language–Action Models}
VLA models enable direct mapping from visual observations and language instructions to robotic actions and have become a dominant paradigm for general-purpose robot control~\cite{ma1778survey,zhong2025survey}. End-to-end VLA models directly generate low-level actions from multi-modal inputs via large-scale pretraining~\cite{brohan2023,kim2024openvla,black2024,intelligence2025}, but suffer from performance degradation on long-horizon tasks due to goal ambiguity and error accumulation~\cite{yang2025lohovla}. Hierarchical methods mitigate this issue by introducing intermediate guidance~\cite{shao2025}. Prior work explores various guidance such as keypoints~\cite{wu2025momanipvla,yuan2024robopoint}, language~\cite{shi2025}, and visual predictions~\cite{chengoal,zhen20243}. While these approaches improve robot control stability and instruction following, they remain vulnerable to error propagation over time and struggle with long-horizon tasks.

\subsection{Task and Motion Planning}

Task and motion planning (TAMP) integrates symbolic reasoning with motion generation to solve multi-step tasks under physical constraints~\cite{Kaelbling2011-mz}, but often suffers from limited scalability. To address this, prior work has explored imitation learning to accelerate planning~\cite{Silver2021-mv, Dalal2023-cq} and reinforcement learning to enhance adaptability in dynamic environments~\cite{Chitnis2016-pd, Paxton2017-lj}. More recently, LLM-based TAMP has leveraged language priors to guide planning~\cite{pmlr-v205-huang23c, Wang2024-fg, chen-etal-2024-prompt}.
Another line of work attempts to learn planning domains to improve scalability~\cite{Diehl2021AutomatedGO, pmlr-v229-kumar23a, Silver2023-mi, Liang2024-hf, mao2023learning, Huang2024-it, Huang2025-ue, han2024interpret}, but primarily focuses on abstract world modeling without visual grounding. A framework that jointly captures logical and visual dynamics to enable simultaneous prediction of logical transitions and visual observations remains lacking.

\section{Preliminary}

\subsection{Symbolic Task Formalization}

We represent the logic-level structure of a robot manipulation task using a PDDL-style symbolic formulation.
A symbolic domain is defined as
\(
\mathcal{D}=(\mathcal{P},\mathcal{A}),
\)
where \(\mathcal{P}\) is a finite set of predicate symbols and \(\mathcal{A}\) is a set of parameterized action schemas.
Given the objects
\(
\mathcal{O}=\{o_1,\ldots,o_n\}
\)
in a task instance, grounding the predicates in \(\mathcal{P}\) with compatible objects produces the set of ground atoms
\(
\mathcal{G}
=
\left\{
p(o_{i_1},\ldots,o_{i_k})
\;\middle|\;
p\in\mathcal{P},\;
o_{i_j}\in\mathcal{O}
\right\},
\)
where \(k\) is the arity of predicate \(p\).
A logical state is represented by the subset of ground atoms that are true,
\(
\mathcal{X}\subseteq\mathcal{G}.
\)
Similarly, grounding an action schema in \(\mathcal{A}\) produces a logical action \(a\), which describes a semantic robot subtask such as picking an object, placing it into a receptacle, or closing a drawer.

A long-horizon task is specified by an initial logical state \(\mathcal{X}_0\) and a goal logical state \(\mathcal{X}_G\subseteq\mathcal{G}\).
We decompose its execution into a sequence of \(M\) symbolic transitions,
\(
\mathcal{X}_0
\xrightarrow{a_1}
\mathcal{X}_1
\xrightarrow{a_2}
\cdots
\xrightarrow{a_M}
\mathcal{X}_M,
\)
where \(m\) indexes symbolic subtask boundaries.
At boundary \(m\), \(\mathcal{X}_m\) denotes the current logical state, \(a_{m+1}\) denotes the next logical action, and \(\mathcal{X}_{m+1}\) denotes the state expected after successfully completing that action.
The corresponding one-step transition can therefore be written as
\(
\mathcal{X}_m
\xrightarrow{a_{m+1}}
\mathcal{X}_{m+1}.
\)
A task is completed when the terminal state satisfies the goal, i.e.,
\(
\mathcal{X}_G\subseteq\mathcal{X}_M.
\)

In classical PDDL planning, each action schema is associated with preconditions and add/delete effects, and the successor state is computed using a deterministic transition function.
In our formulation, the transition dynamics are instead learned from annotated robot trajectories.
Given the current state, task goal, and preceding symbolic trajectory, the logical world model predicts candidate pairs
\(
(a_{m+1}^{(i)},\hat{\mathcal{X}}_{m+1}^{(i)})
\),
where \(\hat{\mathcal{X}}_{m+1}^{(i)}\) is the predicted successor state associated with candidate action \(a_{m+1}^{(i)}\).
After candidate evaluation and selection, we use \(a_{m+1}\) and \(\mathcal{X}_{m+1}\) to denote the selected logical action and predicted successor state, respectively.
This convention distinguishes candidate predictions used during symbolic search from the selected transition provided to the visual world model and VLA.

\subsection{Visual Subgoal Representation}

The symbolic transition specifies what semantic change should occur but does not directly describe the visual and geometric configuration that the robot should reach.
We therefore associate each symbolic subtask with a latent visual subgoal.
At the beginning of subtask \(m+1\), the robot receives an RGB observation \(obs_m\) and its current configuration \(q_m\).
The selected logical action \(a_{m+1}\) describes the subtask to execute, while \(\mathcal{X}_{m+1}\) represents its predicted resulting logical state.
Together, these variables define the conditioning context
$
c_m
=
\left(
obs_m,\,
q_m,\,
a_{m+1},\,
\mathcal{X}_{m+1}
\right).
$
Let
\(
\phi:\mathcal{I}\rightarrow\mathcal{F}
\)
be a fixed visual encoder that maps an RGB image from the image space \(\mathcal{I}\) into a \(d\)-dimensional latent feature space
\(
\mathcal{F}\subseteq\mathbb{R}^{d}.
\)
For each training subtask, \(obs_{\mathrm{goal}}^{m+1}\) denotes the terminal keyframe observed after the successful completion of \(a_{m+1}\).
Its latent representation,
\(
f_{\mathrm{goal}}^{m+1}
=
\phi\!\left(obs_{\mathrm{goal}}^{m+1}\right)
\in\mathcal{F},
\)
serves as the ground-truth visual subgoal.

The visual world model predicts the corresponding latent visual subgoal directly from the current multimodal context,
\[
f_{\mathrm{pred}}^{m+1}
=
M_V
\left(
obs_m,\,
q_m,\,
a_{m+1},\,
\mathcal{X}_{m+1}
\right)
\in\mathcal{F},
\]
where \(M_V\) denotes the visual world model.
Rather than predicting every intermediate observation within the subtask, \(M_V\) predicts only the latent representation of its terminal visual state.
Consequently, \(f_{\mathrm{pred}}^{m+1}\) provides a fixed perceptual target for low-level control throughout the execution of \(a_{m+1}\).
The framework therefore operates at two temporal resolutions.
The subtask index \(m\) is used by the logical and visual world models, which are invoked once at each symbolic transition.
The control index \(t\) denotes the higher-frequency VLA execution steps within the active subtask.
Accordingly, \(obs_m\) and \(q_m\) refer to the observation and robot configuration at a subtask boundary, whereas \(obs_t\) and \(q_t\) denote their control-step counterparts.
During execution of \(a_{m+1}\), the VLA receives the fixed symbolic action and visual subgoal \((a_{m+1},f_{\mathrm{pred}}^{m+1})\) while continuously updating its low-level actions from \((obs_t,q_t)\).

\section{Methodology}

Building on the symbolic transition and visual-subgoal formulation above, H-WM consists of a logical world model, a logic-conditioned visual world model, and a guided VLA policy.

\subsection{Logical World Model}

The logical world model enables long-horizon reasoning in symbolic space.
Classical TAMP frameworks rely on hand-crafted PDDL transition rules~\cite{Kaelbling2011-mz}, which can become brittle when logical state estimates are incomplete or imperfect.
We instead model symbolic transitions directly from data.

We annotate robot trajectories at subtask boundaries with the current logical state $\mathcal{X}_m$, next logical action $a_{m+1}$, resulting logical state $\mathcal{X}_{m+1}$, observation $obs_m$, and robot configuration $q_m$, forming the LIBERO-Logic dataset with unified transition tuples
$\langle\mathcal{X}_m,a_{m+1},\mathcal{X}_{m+1},obs_m,q_m\rangle$.
We initialize from a planning-enhanced language model that has been post-trained for symbolic planning through structured reasoning supervision and planning-oriented preference optimization~\cite{Huang2025-zo}.
We further fine-tune it on LIBERO-Logic to obtain the logical world model $M_L$, adapting its generic symbolic planning capability to robot manipulation.

During inference, $M_L$ is instantiated with two complementary roles, the generator $M_L^{\mathrm{search}}$ and the evaluator $M_L^{\mathrm{eval}}$.
Given the current state $\mathcal{X}_m$, task goal, and preceding symbolic trajectory, $M_L^{\mathrm{search}}$ predicts candidate action--state transitions
$
\mathcal{C}_m=
\left\{
\left(a_{m+1}^{(i)},\hat{\mathcal{X}}_{m+1}^{(i)}\right)
\right\}_{i=1}^{K},
$
where each candidate jointly specifies the next logical action and its predicted resulting state.
The model therefore performs learned symbolic rollout rather than applying an externally specified deterministic transition function.

To reduce error accumulation across long horizons, we perform beam-style search over the predicted transitions.
At each search depth, the top-$B$ partial trajectories are retained and each is expanded with $K$ candidate transitions.
For a partial trajectory $\tau$, $M_L^{\mathrm{eval}}$ predicts a trajectory-consistency score
$s_{\mathrm{eval}}(\tau)$ that reflects logical validity and compatibility with the task goal.
We additionally compute explicit goal progress from the terminal predicted state,

\[
s_{\mathrm{goal}}(\mathcal{X})
=
\frac{|\mathcal{X}\cap\mathcal{X}_{G}|}{|\mathcal{X}_{G}|},
\]
where $\mathcal{X}_{G}$ denotes the goal logical state.
Candidate trajectories are ranked by
$
S(\tau)
=
\lambda_{\mathrm{eval}}s_{\mathrm{eval}}(\tau)
+
\lambda_{\mathrm{goal}}s_{\mathrm{goal}}(\mathcal{X}),
$
and the top-$B$ trajectories are retained for the next expansion.
Search terminates when the predicted state satisfies the goal or a maximum symbolic depth is reached.
In this way, $M_L$ acts jointly as a learned symbolic transition model and a structured trajectory evaluator.

\subsection{Visual World Model}

The visual world model provides visually grounded guidance by predicting the latent representation of the next visual subgoal, conditioned on the current observation, predicted logical state, and corresponding logical action.
Unlike autoregressive visual world models, we predict only the terminal visual representation of each subtask, avoiding long visual rollouts and reducing error accumulation.

As shown in Fig.~\ref{fig:Framework}, the visual world model adopts a multi-expert architecture comprising an understanding expert and a prediction expert.
The understanding expert adopts the PaliGemma VLM backbone of $\pi_{0.5}$~\cite{intelligence2025}.
We freeze its visual encoder and multimodal projector to extract features from $obs_m$, while Gemma jointly processes these features and the text embeddings of $a_{m+1}$ and $\mathcal{X}_{m+1}$.
The prediction expert is a randomly initialized lightweight transformer.
Through unidirectional fused self-attention, its prediction tokens attend to the understanding expert's representations, with $q_m$, noisy subgoal features, and flow time embedded into the prediction stream.
A trainable head $R_\theta$ projects its output into a velocity field in PaliGemma's projected visual feature space.

During training, the terminal keyframe $obs_{\mathrm{goal}}^{m+1}$ is encoded as
$
f_{\mathrm{goal}}^{m+1}=\phi(obs_{\mathrm{goal}}^{m+1}),
$
where $\phi$ includes the frozen visual encoder and multimodal projector.
We train the prediction expert using the conditional flow-matching objective~\cite{intelligence2025, lipman2023flow}, with terminal visual features replacing action targets.
At inference, we integrate the predicted velocity field from Gaussian noise to obtain the visual subgoal $f_{\mathrm{pred}}^{m+1}$.

\subsection{Injecting Hierarchical World Model Guidance in VLA}

The guided VLA extends the $\pi_{0.5}$ with an additional goal expert for incorporating the visual subgoal predicted by the visual world model.
It retains the original understanding and action experts for multimodal state understanding and low-level action generation, while the goal expert provides explicit conditioning on the predicted future visual state.

At control step $t$, the understanding expert combines the current observation $obs_t$ and logical action $a_{m+1}$ to produce the current-state representation $h_t^u$.
In parallel, the goal expert encodes the fixed latent visual subgoal $f_{\mathrm{pred}}^{m+1}$ into a goal representation $h_m^g$.
The action expert takes the current robot configuration $q_t$ and noisy action tokens as inputs and generates an action chunk $\hat{\alpha}{m}^{t:t+k}$ conditioned on both $h_t^u$ and $h_m^g$.
To integrate the three experts, we extend the base VLA's directional expert interaction with an additional goal-to-action pathway.
Specifically, the action expert attends to the hidden representations of both the understanding and goal experts through asymmetric cross-attention.
Information flows only from the understanding and goal experts to the action expert; reverse attention is disabled.
This design retains the base VLA's action-generation mechanism while incorporating both logical and visual guidance.
The guided VLA is trained end-to-end using the same conditional flow-matching objective for continuous action generation as $\pi_{0.5}$~\cite{intelligence2025}.

\subsection{Subtask Completion and Transition Prediction}

During execution, the VLA performs the subtask specified by $a_{m+1}$ toward its predicted visual subgoal.
To detect completion, we use a separate predictor initialized from the understanding expert of the trained guided VLA and augmented with a lightweight classification head.
At each control step $t$, a learnable \texttt{[CLS]} token is processed together with the observation tokens from $obs_t$ and the logical-action tokens from $a_{m+1}$.
Through self-attention, this token gathers information from both inputs.
Its final hidden representation $h_t^{\mathrm{cls}}$ is passed to the classification head to predict the completion probability,
$
p_t^{\mathrm{done}}
=
\sigma(W h_t^{\mathrm{cls}}),
$
where $\sigma$ denotes the sigmoid function.
The predictor is fine-tuned using binary cross-entropy loss and evaluated at every VLA control step.
Once completion is detected, execution advances to the next symbolic transition, and the visual world model predicts the next visual subgoal.

\begin{table*}[t]
\centering
\caption{Performance on the LIBERO-LoHo benchmark. 
For each task, we report Q-Score and Success Rate (in \%). 
The best results are shown in \textbf{bold}, and the second-best results are \underline{underlined}.}
\label{tab:loho}
\setlength{\tabcolsep}{3pt}
\renewcommand{\arraystretch}{1.15}
\small
\begin{tabularx}{\textwidth}{@{} l *{5}{Y} Y @{}}
\toprule
\multirow[c]{2}{*}{\textbf{Methods}}
& \multicolumn{5}{c}{\textbf{Tasks Q-score / Task Success Rate}}\\
\cmidrule(lr){2-7}
& Task 1 & Task 2 & Task 3 & Task 4 & Task 5 & \textbf{Average} \\
\midrule

H-WM-$\pi_{0.5}$ (ours)        
& \textbf{98.0/94.0} 
& \textbf{86.7/60.0} 
& \textbf{74.0/46.0} 
& \underline{70.7}/\textbf{42.0} 
& \textbf{95.0/82.0}
& \textbf{84.9/64.8} \\

H-WM-Stable-Diffusion-$\pi_{0.5}$     
& 92.7/82.0 
& 80.0/46.0
& 65.3/\underline{30.0}
& \textbf{74.0}/\underline{34.0}
& \underline{93.5/80.0}
& \underline{81.1/54.4} \\

Logic-guided $\pi_{0.5}$      
& \underline{95.3/86.0} 
& \underline{84.7/58.0} 
& 54.7/16.0 
& 39.3/{4.0}  
& {92.0/78.0}
& 73.2/48.4 \\

LLM-guided $\pi_{0.5}$   
& 84.7/54.0 
& 80.7/42.0 
& \underline{68.0}/24.0
& {41.3/4.0} 
& 59.5/10.0
& 66.8/26.8 \\

$\pi_{0.5}$                   
& 66.0/4.0  
& 73.3/24.0 
& 54.7/4.0  
& 44.7/0.0  
& 38.0/0.0
& 55.3/6.4  \\

$\pi_{0}$                     
& 54.0/0.0  
& 62.0/28.0 
& 44.0/0.0  
& 31.3/0.0  
& 34.0/0.0
& 45.1/5.6  \\

X-VLA
& 48.8/0.0 
& 16.7/0.0
& 23.3/0.0 
& 33.3/0.0
& 40.0/0.0
& 32.4/0.0 \\

Gr00t-N1.5
& 0.0/0.0 
& 30.7/0.0 
& 34.7/0.0 
& 2.7/0.0 
& 0.0/0.0
& 13.6/0.0 \\

OpenVLA-OFT
& 0.0/0.0 
& 60.0/0.0 
& 17.3/0.0 
& 22.7/0.0 
& 0.0/0.0
& 20.0/0.0 \\

OpenVLA
& 0.0/0.0 
& 2.7/0.0 
& 24.0/0.0 
& 2.7/0.0 
& 0.0/0.0
& 5.9/0.0 \\
\bottomrule
\end{tabularx}
\end{table*}

\section{Experiments}

\subsection{Training Dataset}

We train our hierarchical world model on two datasets: LIBERO-Logic, which augments the LIBERO dataset~\cite{liu2023libero} with temporally aligned logical annotations at every timestep, and RoboCerebra~\cite{han2025robocerebra}. For LIBERO-Logic, annotations are generated through a two-stage process. We first replay each episode and use predefined predicate classifiers to infer logical states, identifying logical actions from the corresponding state changes. We then manually inspect the annotations and correct labeling errors. From these frame-level annotations, we extract subtask-boundary transition tuples \(\langle \mathcal{X}_m, a_{m+1}, \mathcal{X}_{m+1}, obs_m, q_m\rangle\), where \(m\) indexes symbolic subtask boundaries. For RoboCerebra, we directly adopt the provided task decompositions and standardize the logical actions and predicates across tasks to obtain a consistent symbolic representation.

\subsection{Benchmark and Evaluation}
We evaluate H-WM alongside a range of VLA baselines across three long-horizon planning and control benchmarks, including LIBERO-10~\cite{liu2023libero}, RoboCerebra~\cite{han2025robocerebra}, and the customized LIBERO-LoHo. We introduce \textbf{LIBERO-LoHo}, a more challenging long-horizon benchmark derived from LIBERO. It contains five tasks with extended manipulation sequences, constructed by increasing the number of manipulated objects and inter-subtask dependencies relative to the original LIBERO tasks. Specifically, \textbf{Task 1} is a five-step task that places the front butter and chocolate pudding into the top drawer of a wooden cabinet before closing it. \textbf{Task 2} requires placing alphabet soup, butter, and tomato sauce into a basket over six steps. \textbf{Task 3} is a six-step task that places alphabet soup, cream cheese, and butter into a wooden tray. \textbf{Task 4} requires placing the left black bowl, salad dressing, and chocolate pudding into a wooden tray over six steps. \textbf{Task 5} is a seven-step task that places the back butter and chocolate pudding into the top cabinet drawer, closes the drawer, and then places the black bowl on top of the cabinet. Compared with the original LIBERO benchmark, which typically contains 1--3-step tasks, LIBERO-LoHo extends the task horizon to as many as seven steps, enabling a more rigorous evaluation of long-horizon policy performance.
We primarily report \textbf{Success Rate}, defined as the percentage of tasks completed successfully. For extra-long-horizon benchmarks, including LIBERO-LoHo \textit{(up to 7 steps)} and RoboCerebra \textit{(up to 20 steps)}, we additionally report \textbf{Q-Score}, defined as the fraction of completed subgoals among all required subgoals. This metric provides a more informative measure of task progress when full task completion remains difficult.

\begin{figure*}[t]
    \centering
    \includegraphics[width=1.0\textwidth]{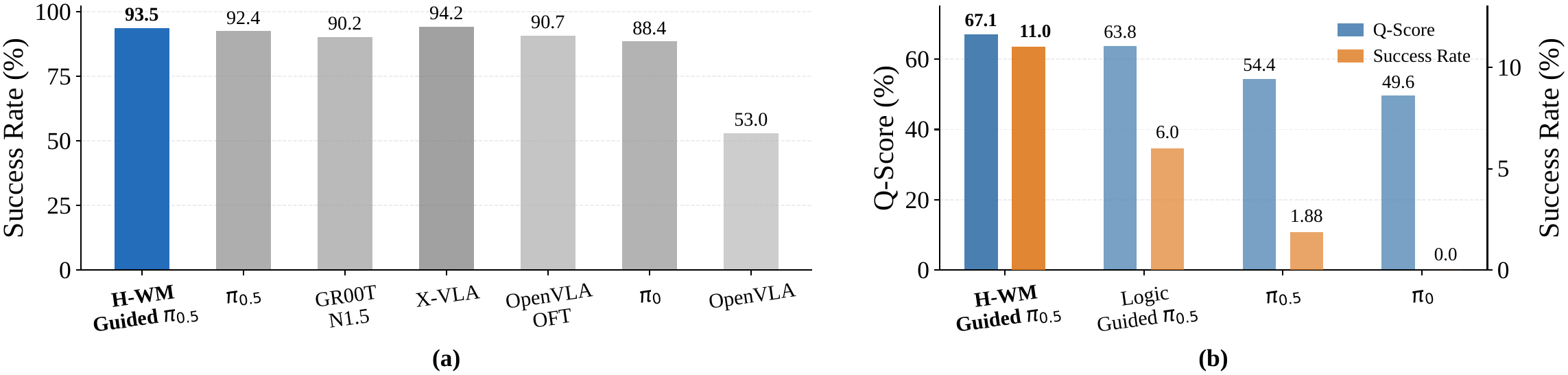}
    \caption{(a) Evaluation results of H-WM-guided $\pi_{0.5}$ and various VLA baselines on the LIBERO-10 benchmark.
    (b) Evaluation results of H-WM-guided $\pi_{0.5}$ and baseline methods on the RoboCerebra benchmark.
    }
    \label{fig3}
    \vspace{-4mm}
\end{figure*}

    
\subsection{Baselines}
We evaluate H-WM against two categories of baselines covering both end-to-end VLA learning and hierarchical long-horizon control. For end-to-end VLAs, we compare with $\pi_0$~\cite{black2024}, $\pi_{0.5}$~\cite{intelligence2025}, OpenVLA~\cite{kim2024openvla}, OpenVLA-OFT~\cite{kim2025fine}, X-VLA~\cite{zheng2026xvla}, and GR00T~\cite{bjorck2025gr00t}. For hierarchical approaches, we include LLM-guided $\pi_{0.5}$, which uses a language model to decompose the task into natural-language subgoals~\cite{shi2025} that are sequentially executed by $\pi_{0.5}$, and Logic-guided $\pi_{0.5}$, which decomposes the task into structured PDDL-style logical actions and states and sequentially conditions $\pi_{0.5}$ on the resulting logical actions without visual-subgoal prediction. These comparisons separate the benefit of hierarchical decomposition from the additional benefit of structured logical guidance and logic-conditioned visual guidance.

\subsection{Implementation Details}

For LIBERO-10 and LIBERO-LoHo, all VLA models are trained on the corresponding LIBERO demonstrations.
For RoboCerebra, models are trained on the provided training split and evaluated on the ideal test set.
The guided VLA is initialized from $\pi_{0.5}$, with the additional goal expert initialized from the Gemma backbone of its understanding expert. We follow its standard training configuration, with a learning rate of $5\times10^{-5}$ and 256 global batch size for 60k steps in PyTorch. The subtask completion predictor is initialized from the PaliGemma backbone of the trained guided VLA, and trained for 10k steps using binary cross-entropy loss with the classification head.
Baseline methods use publicly released LIBERO fine-tuned checkpoints when available.

The logical world model is initialized from the planning-enhanced Qwen3-4B checkpoint released in~\cite{Huang2025-zo} and further full-parameter fine-tuned on LIBERO-Logic for 10k optimization steps with a learning rate of $1\times10^{-5}$ and batch size of 16. We follow the training formulation in~\cite{Huang2025-zo}, including the structured planning data format, autoregressive next-token objective, and optimization setup, while replacing the original planning data with LIBERO-Logic trajectories. For symbolic search, we use a beam width of $B=3$, generate $K=2$ candidate transitions per expansion, set $\lambda_{\mathrm{eval}}=0.5$ and $\lambda_{\mathrm{goal}}=0.5$, and limit the maximum search depth to 20 symbolic steps.

For the visual world model, the visual encoder is instantiated with SigLIP~\cite{zhai2023}, while the multimodal transformer is initialized from the Gemma understanding expert of $\pi_{0.5}$.
The prediction expert is a randomly initialized lightweight Transformer of approximately 300M parameters following the $\pi_{0.5}$ expert architecture.
The projection head uses two fully connected layers with dimensions $1024 \rightarrow 4096 \rightarrow 2048$.
The visual world model is trained for 100k steps with a learning rate of $5\times10^{-5}$ and 256 global batch size.

\subsection{Ablation Studies}
To validate the contribution of visual guidance beyond logical reasoning, we use Logic-guided $\pi_{0.5}$ as a controlled ablation in which the visual world model is removed and only logical guidance is retained. In addition, we replace latent visual prediction with pixel-level terminal-subgoal generation using Stable Diffusion v1.5~\cite{rombach2022high}. The diffusion model is fine-tuned on the same subtask-aligned training data, conditioned on the current observation and predicted logical transition, and its generated subgoal image is encoded by the same frozen visual encoder before being provided to the guided VLA for execution. This compares the effectiveness of latent versus pixel-level visual guidance.

\begin{figure*}[tb]
    \centering
    \includegraphics[width=\textwidth]{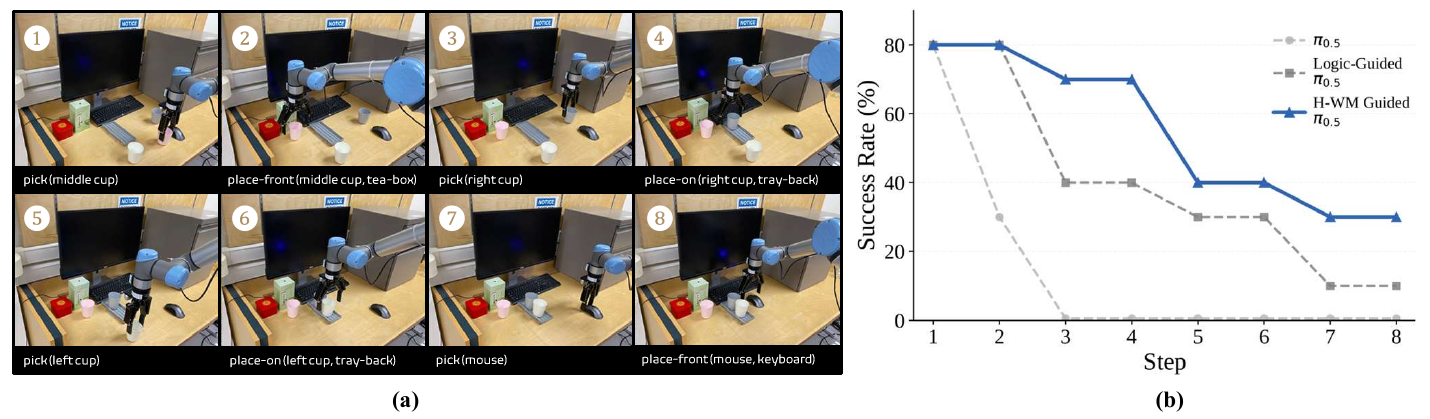}
    \caption{(a) Real-world experiment with UR5e robot evaluating H-WM guided $\pi_{0.5}$ on long-horizon manipulation task. (b) Step-wise success rates in real-robot experiments.}
    \label{fig:real_robot}
\end{figure*}

\subsection{Real-Robot Experiment.}
We evaluate H-WM on a UR5e robot with visual observations captured by an Intel RealSense camera. The robot performs an eight-step office-table cleaning task involving multiple sequential pick-and-place operations, which requires both spatial reasoning and consistent task progression over a long horizon. We compare $\pi_{0.5}$, logic-guided-$\pi_{0.5}$, and H-WM-guided $\pi_{0.5}$ under the same setup. All methods are trained using the same 40 manually collected demonstration trajectories and executed through the UR5e control API. Each method is evaluated over 20 real-robot trials, and performance is measured using step-wise success rate to capture both partial task progress and complete long-horizon execution. Logical states are estimated from visual observations using Qwen3-VL-2B with one-shot prompting.

\section{Results}

Our experiments aim to answer the following questions: (1) Does H-WM provide effective guidance for downstream VLAs? (2) How does the H-WM-guided VLA compare to baseline approaches and models? (3) How does each level of the world model contribute to the final performance? (4) Is the H-WM framework deployable in real-robot settings?

\textbf{(1) Effectiveness of H-WM guidance:} The results demonstrate that H-WM provides stable and effective guidance for long-horizon tasks across three benchmarks. On LIBERO-LoHo (\Cref{tab:loho}), H-WM-guided $\pi_{0.5}$ significantly outperforms $\pi_{0.5}$, improving success rate by 58.4 percentage points and Q-score by 29.6 points. Similar gains are observed on RoboCerebra (\Cref{fig3}.(b)), with improvements of 12.7 points in Q-score and 9.1 percentage points in success rate.
In contrast, $\pi_{0.5}$ attains relatively high Q-Scores but low Success Rates, often due to missing intermediate steps, ignored instructions, and incorrect action ordering, leading to partial completion but overall failure.
On LIBERO-10 (\Cref{fig3}.(a)), H-WM guidance still improves performance, though the margin is smaller. This is because LIBERO-10 tasks are much shorter, allowing VLAs to succeed directly.

\textbf{(2) Comparison to Baselines:} H-WM-guided $\pi_{0.5}$ consistently outperforms all end-to-end VLA baselines on long-horizon tasks (\Cref{tab:loho}, \Cref{fig3}.(b)). Despite their strong low-level manipulation capabilities, end-to-end VLAs struggle to maintain task progress over extended horizons. Without explicit intermediate guidance, they are sensitive to prompt variations and frequently omit required subtasks, execute them in the wrong order, or terminate before satisfying the complete task goal. H-WM mitigates these failures by explicitly predicting and executing a sequence of intermediate transitions. We further compare H-WM with hierarchical methods that decompose the original instruction before VLA execution. LLM-guided $\pi_{0.5}$ improves over raw VLAs through natural-language task decomposition, but its language subgoals remain ambiguous and do not explicitly track logical state changes. Logic-guided $\pi_{0.5}$ introduces structured PDDL-based actions and states, improving task ordering and global consistency, but provides no perceptual description of the desired outcome of each subtask. H-WM combines the complementary strengths of both levels: logical transitions maintain globally consistent task progression, while latent visual subgoals provide perceptually grounded targets for low-level control. This joint guidance produces more reliable long-horizon execution than either end-to-end policies or language- and logic-only hierarchical methods. On the shorter-horizon LIBERO-10 benchmark (\Cref{fig3}.(a)), intermediate guidance is less critical; consequently, H-WM shows smaller gains and ranks second overall.

\textbf{(3) Ablation Study:} The ablation results further validate the necessity of bilevel guidance. As shown in \Cref{tab:loho}, logic-only guidance already surpasses the unguided baseline by 42.0 percentage points in success rate and 17.9 points in Q-score. Incorporating visual guidance yields a further 16.4 percentage-point improvement in success rate and 11.7 points in Q-score. This confirms that visual world modeling provides actionable grounding for symbolic plans, improving alignment between symbolic constraints and perceptual execution.  The results on RoboCerebra (\Cref{fig3}.(b)) support the same conclusion.
Moreover, the results show that pixel-level image generation is less effective than latent visual features. The H-WM-Stable-Diffusion-$\pi_{0.5}$ variant replaces latent visual prediction with Stable Diffusion-based image generation, yet achieves smaller performance gains than the default H-WM, exhibiting a 10.4\% drop in overall success rate. This is likely because pixel-level generation introduces unnecessary visual details and reconstruction noise, whereas latent features provide more compact guidance for VLAs.

\textbf{(4) Real-World Experiment:}
The complete real-robot task is illustrated in \Cref{fig:real_robot}(a), and the corresponding step-wise success rates are reported in \Cref{fig:real_robot}(b). Although $\pi_{0.5}$ can complete early manipulation steps, its success rate declines rapidly as execution progresses. Logic-guided $\pi_{0.5}$ improves task progression through structured subtask guidance, but its performance remains limited in later stages, suggesting that logical guidance alone is insufficient for reliable execution. H-WM complements this logical guidance with predicted visual subgoals, providing explicit perceptual targets for each pick-and-place operation. This combination helps connect the intended task progression with the physical configuration required at each stage. Consequently, H-WM-guided $\pi_{0.5}$ maintains higher success rates throughout the later steps and achieves more reliable full-task completion, demonstrating the effectiveness of hierarchical world-model guidance in real-world manipulation.




\section{Conclusion}

We propose a hierarchical world model (H-WM) that jointly models logical and visual world dynamics to provide bilevel guidance for VLA models in long-horizon robotic tasks. The logical world model captures global task structure and long-term dependencies, while the visual level grounds logical transitions into visually meaningful latent subgoals, enabling stable guidance of VLAs over extended horizons.
Experimental results show that H-WM significantly improves long-horizon performance over base VLA models and hierarchical or end-to-end approaches. These findings demonstrate that hierarchical world modeling offers effective and scalable guidance for bridging symbolic reasoning and perceptual grounding in VLA systems. Although the proposed Hierarchical World Model demonstrates strong performance in guiding long-horizon manipulation tasks, several limitations remain. 
H-WM introduces additional training cost and system complexity and assumes tasks admit meaningful symbolic representations. Future work will explore self-supervised learning to reduce logical supervision and incorporate additional sensory modalities for spatial reasoning.

\bibliographystyle{IEEEtran}
\bibliography{refs}

\end{document}